\pdfoutput=1

\documentclass[11pt]{article}

\usepackage{acl}

\usepackage{times}
\usepackage{latexsym}

\usepackage[T1]{fontenc}

\usepackage[utf8]{inputenc}

\usepackage{microtype}

\usepackage{hyperref}       
\usepackage{url}            
\usepackage{booktabs}       
\usepackage{nicefrac}       
\usepackage{xcolor}         
\usepackage{multirow}
\usepackage{bm}

\usepackage{times}
\usepackage{latexsym}

\usepackage{amsmath}

\usepackage{algorithm}
\usepackage{algpseudocode}
\usepackage{graphicx}  
\usepackage{CJK}
\usepackage{bbm}
\usepackage{float}
\usepackage{xcolor,colortbl}

\usepackage{bigstrut,bigdelim}
\usepackage{paralist}
\usepackage{diagbox}
\usepackage{wrapfig}
\usepackage{verbatim}

\usepackage{subfigure}
\usepackage{multicol}
\usepackage{multirow}
\usepackage{makecell}
\usepackage{amssymb}
\usepackage{xcolor}
\newcommand{\thickhline}{\noalign{\hrule height 1pt}}

\title{Learning to Express in Knowledge-Grounded Conversation}

\author{\\
    \textbf{Xueliang Zhao\textsuperscript{1,2}, Tingchen Fu\textsuperscript{3}, Chongyang Tao\textsuperscript{4}, Wei Wu\textsuperscript{5}, Dongyan Zhao\textsuperscript{1,2}\footnotemark[1], Rui Yan\textsuperscript{3}\footnotemark[1]}\\
    \textsuperscript{1}Wangxuan Institute of Computer Technology, Peking University\\
    \textsuperscript{2}Center for Data Science, AAIS, Peking University\\
    \textsuperscript{3}Gaoling School of Artificial Intelligence, Renmin University of China\\
    \textsuperscript{4}Microsoft Corporation \quad
    \textsuperscript{5}Meituan, Beijing, China\\
    \texttt{\{xl.zhao,zhaody\}@pku.edu.cn} \quad \texttt{ruiyan@ruc.edu.cn} \\
    \texttt{\{lucas.futingchen,chongyangtao,wuwei19850318\}@gmail.com}\\
}

\begin{document}
\maketitle

\renewcommand{\thefootnote}{\fnsymbol{footnote}}
\footnotetext[1]{Corresponding authors: Dongyan Zhao and Rui Yan.}
\setcounter{footnote}{0}
\renewcommand{\thefootnote}{\arabic{footnote}}

\begin{abstract}
Grounding dialogue generation by extra knowledge has shown great potentials towards building a system capable of replying with knowledgeable and engaging responses. Existing studies focus on how to synthesize a response with proper knowledge, yet neglect that the same knowledge could be expressed differently by speakers even under the same context. In this work, we mainly consider two aspects of knowledge expression, namely the structure of the response and style of the content in each part. We therefore introduce two sequential latent variables to represent the structure and the content style respectively. We propose a segmentation-based generation model and optimize the model by a variational approach to discover the underlying pattern of knowledge expression in a response. Evaluation results on two benchmarks indicate that our model can learn the structure style defined by a few examples and generate responses in desired content style.

\end{abstract}

\section{Introduction}

Building an open domain dialogue system has attracted increasing attention from the community of AI and NLP. 
Despite the impressive progress, existing models are notorious for replying with generic and bland responses. To bridge the gap, researchers resort to ground dialogue generation by extra knowledge such as unstructured documents \cite{zhou2018dataset,dinan2018wizard}. By this means, the documents serve as content sources and make a dialogue system knowledgeable regarding various concepts in a discussion.

However, existing studies focus on how to synthesize a response with proper knowledge \cite{dinan2018wizard,kim2020sequential,zhao2020knowledge}, but pay little attention to the fact that the same knowledge could be expressed differently even under the same context. These models usually employ a regular decoder to generate the response in an auto-regressive manner given the contextual representations of knowledge and dialogue context, which makes the generation process less explainable and controllable.

\begin{table}[]
\centering
\includegraphics[width=1.0\linewidth]{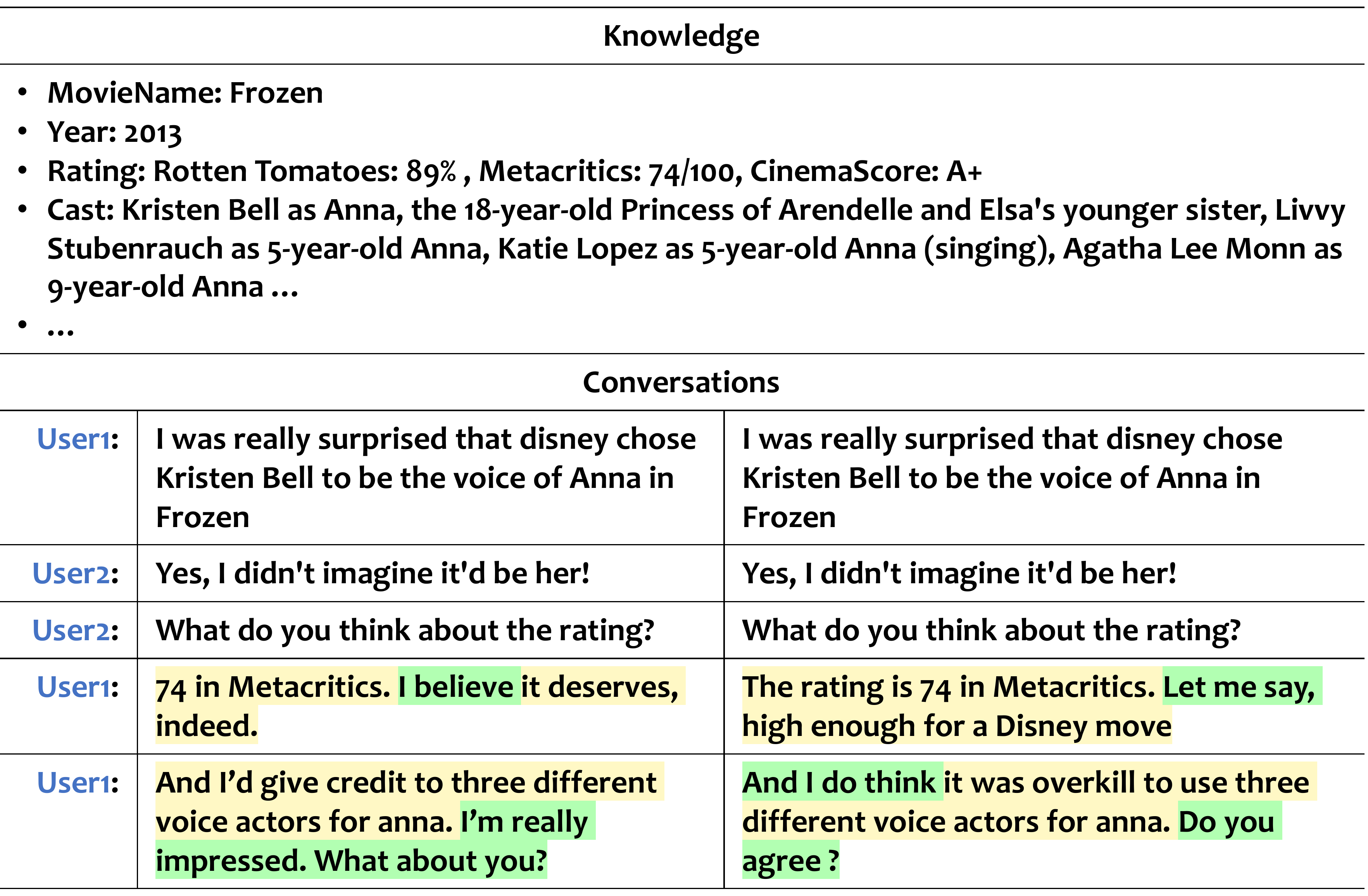}
\caption{A case from CMU$\_$DoG. Given the same knowledge and context, the last two turns in left and right conversations exhibit positive and negative sentiments, respectively. Each utterance can be decomposed into \colorbox{yellow!30}{knowledge-related} and \colorbox{green!30}{knowledge-irrelevant} segments. }
\label{tab:intro_case}
\vspace{-6mm}
\end{table}

In general, the expression style of response is composed of two aspects: the structure of the response and the style of the content in each part. 
As the example shown in Table \ref{tab:intro_case}, knowledge-related phrases and clauses tend to be long, like ``And I'd give credit to three different voice actors for anna.'', or short, like ``74 in Metacritics''. Besides, they may appear at the beginning of the sentence, or at the end.
For the sake of description, we decompose a response into a sequence of non-overlapping segments, each is either related to certain background knowledge and diverse in content style, or almost irrelevant to the knowledge but simply playing the role of stitching the context and carrying on the conversation. 
We therefore define the structure style as the distribution and number of two kinds of segments. The structure style itself is far from dominant in the sentence expression, since different speakers could convey converse attitude even if the context and the knowledge are exactly the same. So it is necessary to introduce the content style as the expression fashion within each knowledge-related segment. 
We further introduce two latent variables to facilitate end-to-end training, one for predicting the start and end positions of a segment, the other for deciding the category of each segment. 
Since the human annotations for sentence segmentation are absent and enumerating over all possibilities to maximize the likelihood of the response is time-consuming, we propose a variational framework for segmentation-based generation and induce an evidence lower bound of the likelihood.

Formally, our model follows an encoder-decoder architecture. The encoder is to obtain the contextual representation of conversational context and knowledge in a regular way. The decoder consists of three types of modules: (1) context module, for response only based on context without knowledge; (2) plain-knowledge module, for response referring knowledges but without particular style; and (3) stylized-knowledge module, for response referring knowledges and with a specific style. The context module is the only module not relying on knowledge, but simply paying attention to contextual information. Compared with plain-knowledge module, stylized-knowledge module has unique adapters, which is their primary discrepancy. 
When decoding, the decoder first predicts the segmentation of the response and then makes a choice in three kinds of modules to generate a single segment. Both the segmentation and the module selection are instructed under sequential latent variables.

We train our model on the Reddit Corpus published by \cite{li2020zero} and evaluate our model on two benchmarks of knowledge-grounded conversation: Wizard of Wikipedia(Wizard) \cite{dinan2018wizard} and CMU Document Grounded Conversation(CMU$\_$DoG) \cite{zhou2018dataset}.
Evaluation results indicate that our model can significantly outperform state-of-the-art methods in the zero-resource setting (i.e., only trained on the Reddit Corpus).
In addition, the performance of our model improves significantly on Wizard and CMU$\_$DoG with the presence of only $10\%$ training data and the segment distributions after fine-tuning are consistent with our prior knowledge about the two datasets, indicating that our model can learn the structure style with little cost. 
Finally, our model outperforms previous state-of-the-art models on the accuracy of performing sentiment classification using generated responses,
which indicates that the model can be controlled to express knowledge with the desired content style.

Contributions in this work are three-fold: 
(1) exploration of the knowledge expression in knowledge-grounded conversation; 
(2) proposal of a variational segmentation-based generation model to discover the underlying expression style in a response; 
(3) empirical verification of the effectiveness of the proposed model on two benchmarks of knowledge-grounded conversation.

\section{Related Work}
On the vanilla encoder-decoder architecture~\citep{shangL2015neural,vinyals2015neural}, various extensions have been made to  model the structure of dialogue contexts \cite{serban2016building,serban2017hierarchical,zhang2019recosa}; to improve diversity of responses \cite{li2015diversity,xing2017topic,zhao2017learning,tao2018get};  to control attributes of responses \cite{xu2019neural,zhou2017emotional,wang2018learning,see2019makes}; and to bias responses to some specific personas \cite{li2016persona,zhang2018personalizing}. Recently, grounding dialogue generation by extra knowledge has seemed promising to bridge the gap between conversation with existing systems and conversation with humans, and the knowledge could be obtained from knowledge graphs \cite{zhou2018commonsense,moon2019opendialkg,tuan2019dykgchat}, retrieved from unstructured documents \cite{dinan2018wizard,lian2019learning,zhao2019document,zhao2020low,kim2020sequential,li2020zero,fu2022there} or visual background \cite{mostafazadeh2017image,shuster2018engaging,huber2018emotional}. In this work, we study document-grounded dialogue generation. Rather than selecting knowledge relevant to dialogue context and directly exploiting pre-trained language models to generate the response, we focus on expressing knowledge in this task.
 
The idea of sequence modeling via segmentation \cite{wang2017sequence,kim2019variational} has attracted widespread attention in several natural language processing tasks. In text segmentation task, \citet{wang2017sequence} propose a probabilistic model for sequence modeling via their segmentation and a ``Sleep-WAke Network''(SWAN) method. In machine translation, \citet{huang2017towards} propose a neural phrase-based machine translation system that models phrase structures in the target language using SWAN. In data-to-text generation, \citet{wiseman2018learning} develop a neural template-like generation model with an HSMM decoder, which is learned tractably by backpropagating through a dynamic program; to tackle the problem of weak Markov assumption for the segment transition probability,  \citet{shen2020neural} propose to explicitly segment target text into fragments and align them with their data correspondences, and jointly learn the segmentation and correspondence via dynamic programming.

\begin{figure*}
\centering
\includegraphics[width=0.7\textwidth]{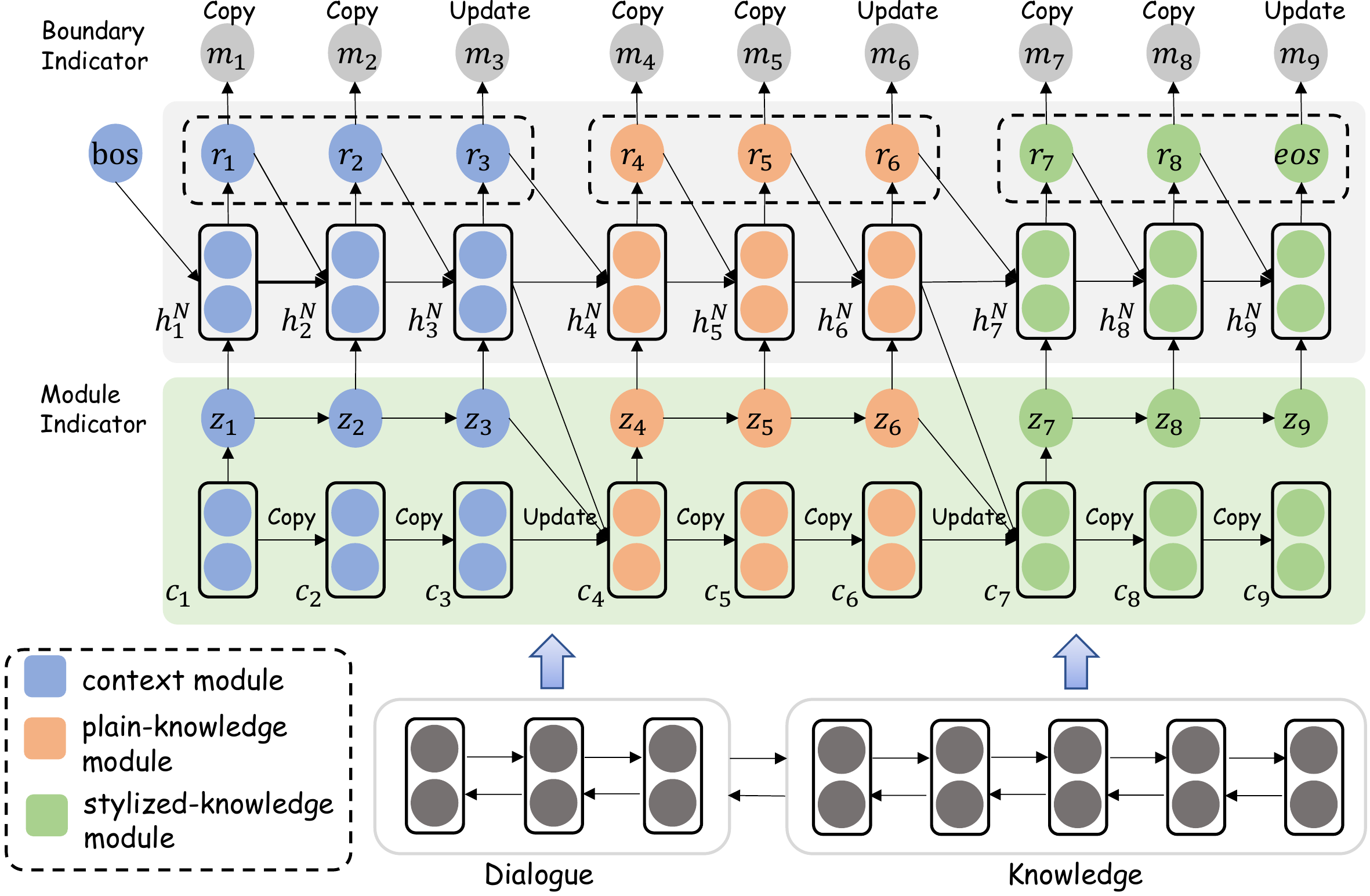}
\caption{Architecture of the proposed model. ``Copy'' means $m_t=0$ and the state of the module indicator $c_{t+1}$ remains unchanged. ``Update'' indicates that $m_t=1$ and the state has been updated to include the information from the previous segment.}
\label{fig:model}
\end{figure*}

\section{Approach}

We start by providing the problem formalization and overview of the proposed model in Sec.\ref{sec:overview}. Then in Sec.\ref{sec:architecture} we describe the design for each components. Lastly, we elaborate how to optimize the components with variational inference and weak supervision in Sec.\ref{sec:optimization}.

\subsection{Overview}
\label{sec:overview}

Suppose that we have a dataset $\mathcal{D} = \{(U_i, K_i, R_i)\}_{i=1}^N$, where $\forall i \in \{1,\ldots,N\}$, $K_i$ serves as background knowledge of the dialogue $(U_i, R_i)$ with $K_{i,j}$ being the $j$-th sentence, $U_i$ is the context of the dialogue with $U_{i,j}$ the $j$-th utterance, and $R_i$ is the response.
To bias the expression to a specific structure style, we further assume that there are a few examples $\mathcal{D}_{sty} = \{(U_i, K_i, R_i)\}_{i=1}^M$ provided by users depicting the required style for knowledge expression. Note that we have $N \gg M$, since corpus in a specific expression style is rare and difficult to acquire.
The goal is to learn a generation model $p_{\theta}(R|U, K)$ ($\theta$ denotes the parameters of the model) from $\mathcal{D}$, to generate a response $R$ following $p_{\theta}(R|U, K)$ given a new dialogue context $U$ and the associated knowledge $K$. Different from previous KGC generation model, we allow users to either (1) bias the structure style of $P_{\theta}(R|U, K)$ to $\mathcal{D}_{sty}$ with little cost; or (2) switch the content style of knowledge expression in $R$.

Figure \ref{fig:model} gives an overview of the proposed model, which is based on the encoder-decoder architecture. The encoder generates the contextual representations of the dialogue and knowledge, while the decoder generates the segments one after another. 
$\mathbf{h}_{t}^{N}$ encodes the dialogue context up to timestep $t-1$ with $N$ denoting the number of decoder layers.
Given $R=(r_1, \cdots, r_t, \cdots, r_{l_r})$ with $r_t$ referring the $t$-th token of $R$ whose length is supposed to be $l_r$, 
the variable $Z=\{z_t\}_{t=1}^{l_r}$ is utilized to control the choice of module of each segment (\textbf{Module Indicator}), and its historical information is encoded by $\{\mathbf{c}_t\}_{t=0}^{l_r}$. 
$M=\{m_t\}_{t=1}^{l_r}$ is a sequence of binary variables and used to determine the boundary of each segment (\textbf{Boundary Indicator}).
Specifically, $m_t=1$ indicating that the current segment is already completed and a new segment should be created at the next timestep. Otherwise, $m_t=0$ and the current segment remains unfinished.
The generative process is disassembled into two steps: (1) determine the type of a new segment based on previously generated text and previous segment types; (2) generate within the current segment until the binary variable $m_t=1$.

\subsection{Model Architecture}
\label{sec:architecture}
\paragraph{Context and Knowledge Encoding.}
We exploit the pre-trained BART~\cite{lewis2020bart} as the backbone of our architecture, which is pre-trained using a variety of denoising objectives and achieves state-of-the-art results on a range of text generation tasks. 
Given the dialogue context $U=(U_1, \cdots, U_n)$, we simply concatenate them as $(u_1, \cdots, u_{l_u})$. Similarly, we concatenate the associated knowledge $K=(K_1, \cdots, K_m)$ as $(k_1, \cdots, k_{l_k})$. $l_u$ and $l_k$ are the length of dialogue context and background knowledge respectively. 
The input of the encoder is then defined as:
\begin{equation} 
    I=[\mathrm{BOS}] k_1 \ldots k_{l_k}[\mathrm{EOS}] u_1 \ldots u_{l_u}[\mathrm{EOS}].
\end{equation}
The input $I$ is truncated or padded to the maximum capacity and then passes through the stacked self-attention layers and results in a knowledge-aware context representation $\mathbf{C}$, and a context-aware knowledge representation $\mathbf{K}$. Specifically, the context-aware knowledge representation is defined as $\mathbf{K}=[\mathbf{h}^{enc}_{1}, \cdots, \mathbf{h}^{enc}_{l_k + 1}]$ where $\mathbf{h}^{enc}_t$ is the last layer of BART encoder at time $t$.
Similarly, the knowledge-aware context representation is defined as $\mathbf{C}=[\mathbf{h}^{enc}_{l_k+2}, \cdots, \mathbf{h}^{enc}_{l_k + l_u + 2}]$.

\paragraph{Prior of Module Indicator.} 
We use the sequential discrete latent variable $Z=\{z_t\}_{t=1}^{l_r}$ to decide which module to invoke at each timestep. The transition of $z_t$ occurs only when a segment is completed, which is decided by the binary boundary variable $M$. 
The prior quantifies the distribution of $z_t$ before we observe the segment, and it is reasonable to assume that the prior of $z_t$ depends on previous module choices $z_{<t}$ and previously generated text. 
Then the transition of $Z$ is defined as:
\begin{equation} 
\begin{aligned}
p_{\theta_{z}}(z_{t}|r_{<t},z_{<t},m_{t-1}) = m_{t-1} \cdot \tilde{p}(z_{t}|\mathbf{c}_{t}) \\
+ (1-m_{t-1}) \cdot \delta(z_{t}=z_{t-1}),
\end{aligned}
\end{equation}
where $\delta$ is a Kronecker delta function. $\mathbf{c}_t$ encodes all previous latent states $z_{<t}$ and generated text $r_{<t}$ as follows:
\begin{equation}
\mathbf{c}_{t} = m_{t-1} \cdot f_{z-\mathrm{rnn}}(\mathbf{\tilde{z}}_{t-1}, \mathbf{c}_{t-1}) + (1-m_{t-1}) \cdot \mathbf{c}_{t-1}.
\end{equation}
$\mathbf{\tilde{z}}_{t-1}=[\mathbf{e}_{t-1};\mathbf{h}_{t-1}^{N, dec}]$ with $\mathbf{e}_{t-1}$ the embedding of $z_{t-1}$ and $\mathbf{h}_{t-1}^{N, dec}$ the representation of last generated token. 
Specifically, $m_{t-1}=0$ means that the next timestep $t$ is still in the same segment as the previous timestep $t-1$ and thus the latent variable $z_t$ should not be updated. 
Otherwise, it means that current segment is completed and $z_t$ is updated with the transition function $\tilde{p}(z_{t}|\mathbf{c}_{t})$.
Because we only have $N_{sty} + 2$ options when choosing a module, where $N_{sty}$ is the number of different user-defined styles in addition to $2$ default styles, so in this model, the latent variable $z_t$ ranges in natural integer to denote corresponding style type.
Specifically, $z_t=0$ denotes choosing the context expression module to generate a knowledge-irrelevant segment; $z_t=1$ tells the model to choose the knowledge expression module without specially customized style; we leave the $z_t \geq 2$ to be user-defined so as to select the knowledge expression module combined with customized style.
The transition function $\tilde{p}(z_{t}|\mathbf{c}_{t})$ is then implemented as a multinomial distribution parameterized by
$\text{Softmax}(f_{z-\mathrm{mlp}}(\textbf{c}_t))$\footnote{We use $f_{*-\mathrm{mlp}}$ to denote a multi-layer perceptron network in this paper.}.

\vspace{-1mm}
\paragraph{Prior of Boundary Indicator.}
The boundary indicator $M=\{m_t\}_{t=1}^{l_r}$ depicts the segmental structure of the response, with $m_{t}=1$ indicates that a new segment should start at time $t+1$. 
Presumably, the prior of $m_t$ could be inferred from $r_{\leq t}$ and $z_t$.
We model the distribution $p_{\theta_{m}}(m_t|r_{\leq t}, z_t)$ by a Bernoulli distribution parameterized by $\sigma(f_{m-\mathrm{mlp}}([\mathbf{e}_{t};\mathbf{h}_{t}^{N, dec}]))$, where $\sigma$ denotes the sigmoid function.

\vspace{-1mm}
\paragraph{Stylized Generation.}
As mentioned above, the generation process involves scheduling different modules according to $z_t$.
Here we give a systematic description of the generation process. 
The decoder accepts the token generated last timestep $r_{t-1}$ as input, performs transformation in $N$ decoder layers, finally obtains a dense representation.

We use $\mathbf{h}^{l}_{t}$ to denote the hidden state after the $l$-th layer at timestep $t$, which is a shorthand for $\mathbf{h}^{l,dec}_{t}$ for brevity.
Specially, $\mathbf{h}^{0}_{t}$ is the output of the embedding layer.
When $z_t=0$, it implies that knowledge encoding is unnecessary for current segment so $\mathbf{h}^{l}_{t}$ is defined as:
\begin{equation}
    \mathbf{h}^{l}_{t}=\text{DecoderLayer}(\mathbf{h}^{l-1}_{t}, \mathbf{H}^{l-1}_{t-1}, \mathbf{C}),
\end{equation}
where $\mathbf{H}^{l}_{t-1}=[\mathbf{h}^{l}_1, \cdots, \mathbf{h}^{l}_{t-1}]$
is a sequence of decoder hidden states in previous timesteps, and $\mathbf{C}$ is the context representation mentioned above. 
$\text{DecoderLayer}(\cdot,\cdot,\cdot)$ is implemented as pre-trained BART decoder layer where $\mathbf{h}^{l-1}_{t}$ first plays self-attention on $\mathbf{H}^{l-1}_{t-1}$ then performs cross-attention on $\mathbf{C}$.
The probability $p(r_t|r_{<t}, z_{t}=0)$ is defined as a multinomial distribution parameterized by $\text{Softmax}(f_{r-\mathrm{mlp}}(\textbf{h}^{N}_{t}))$, where $\mathbf{h}^{N}_{t}$ encodes the generated tokens up to timestep $t-1$.
When $z_t=1$, the implementation of decoder layer is analogous to the $z_t=0$ case except that we replace $\mathbf{C}$ with $\mathbf{K}$, since knowledge is needed:
\begin{equation}
    \mathbf{h}^{l}_{t}=\text{DecoderLayer}(\mathbf{h}^{l-1}_{t}, \mathbf{H}^{l-1}_{t-1}, \mathbf{K}).
\end{equation}

To generate a segment with a particular customized style when $z_t \geq 2$, we introduce some adapters to bias the generation efficiently following \citet{houlsby2019parameter}. Specifically, the hidden state $\mathbf{h}^{l}_{t}$ is defined as:
\begin{equation}
    \mathbf{h}^{l}_{t}=\text{DecoderLayer}_{adp}(\mathbf{h}^{l-1}_{t}, \mathbf{H}^{l-1}_{t-1}, \mathbf{K}),
\end{equation}
where $\text{DecoderLayer}_{adp}(\cdot, \cdot, \cdot)$ denotes the transformer decoder layer with adapters inserted.
To make the style fine-grained and adjustable, each style has a unique set of adapters. Different styles have no adapter in common. In addition, our model has the ability to learn to express in any style, as long as a discriminator for the desired style is provided.\footnote{The proposed method is also able to control the content style of knowledge-irrelevant segmentation by introducing extra adapters. But we focus on knowledge expression in this work.}

\subsection{Learning Details}
\label{sec:optimization}

We introduce auxiliary distributions $q_{\phi_{m}}(M | R)= $ $\prod_{t=1}^{l_r} q_{\phi_{m}}(m_t | R)$ and $q_{\phi_{z}}(Z | M, R)= $ $\prod_{t=1}^{l_r} q_{\phi_{z}}(z_t | M, R)$, which serve as an approximation to the intractable posterior of the boundary indicator $M$ and the module indicator $Z$. 
We then apply variational approximation which gives the following evidence lower bound objective \footnote{We always have $m_0=1$}(ELBO) \cite{hoffman2013stochastic}:
\vspace{-1mm}
\begin{equation}\small
\begin{aligned}
& \log p_{\theta}(R | U, K) \\
& \geq \mathbb{E}_{q_{\phi_{m}}(M | R)} \left(\mathbb{E}_{q_{\phi_{z}}(Z | M, R)} \sum_{t=1}^{l_r} \log p_{\theta}(r_t | r_{<t}, z_t) \right.  \\
& \left. - \sum_{t=1}^{l_r} m_{t-1} \cdot D_{\mathrm{KL}}\big(q_{\phi_{z}}(z_t | M,R) \| p_{\theta_{z}}(z_t)\big) \right) \\
& - \sum_{t=1}^{l_r} D_{\mathrm{KL}}\big(q_{\phi_{m}}(m_t | R) \| p_{\theta_{m}}(m_t)\big),
\end{aligned}
\label{eq:obj}
\vspace{-1mm}
\end{equation}
where $p_{\theta_{z}}(z_t)$ and $p_{\theta_{m}}(m_t)$ stand for $p_{\theta_{z}}(z_{t} | r_{<t},z_{<t},m_{t-1})$ and $p_{\theta_{m}}(m_t | r_{\leq t}, z_t)$ respectively, and $D_{\mathrm{KL}}(\cdot \| \cdot))$ refers to Kullback–Leibler divergence. Detailed derivations are presented in the appendix.

Based on the intuition that the response provides hints about the segmentation, we construct the posterior distribution $q_{\phi_{m}}(m_t | R)$ as a Bernoulli distribution parameterized by $\sigma(f_{m-\mathrm{mlp}}^{\prime}(\psi_{t}))$. $\psi_{t}$ is a feature extracted from a bi-directional LSTM $\psi(R)$. 
Since the module indicator is kept unchanged within a segment, the posterior distribution $q_{\phi_{z}}(z_t | M, R)$ is conditioned on the boundary indicator $m_{t-1}$ and defined as:
\begin{equation} 
    \begin{aligned}
    q_{\phi_{z}}(z_t | M, R) = & m_{t-1}  \cdot \tilde{q}(z_t | \psi_{t}) \\
     & + (1 - m_{t-1}) \cdot \delta(z_{t}=z_{t-1}),
    \end{aligned}
\end{equation}
where $\delta(\cdot)$ denotes Dirac delta function and the transition function $\tilde{q}(z_t | \psi_{t})$ is implemented as a multinomial distribution parameterized by $\text{Softmax}(f_{z-\mathrm{mlp}}^{\prime}(\psi_{t}))$. 
Once we have the posterior distribution, we apply Gumbel-Softmax \cite{jang2016categorical} to take samples of $m_t$ and $z_t$.

\paragraph{Weak Supervision on \textbf{M} and \textbf{Z}.}
We first use StanfordNLP toolkit \cite{manning2014stanford} to parse every response in the training set as a sequence of segments, and use $\tilde{M}=\{\tilde{m}_t\}_{t=1}^{l_r}$ to denote the results of segmentation labeling. 
The pseudo label of module choice $\tilde{Z}=\{z_t\}_{t=1}^{l_r}$ is tagged in a similar way to multiclass classification, determined by (1) the similarity between each segment and knowledge and (2) the classification confidence of the style discriminator. More details about the construction of $\tilde{Z}$ and $\tilde{M}$ are provided in the appendix.

With $\tilde{Z}$ and $\tilde{M}$, the loss function of weak supervision is defined as:
\vspace{-1mm}
\begin{equation} 
\begin{aligned}
\small
\mathcal{L}_{m} &= -\sum_{t=1}^{l_r} \log p_{\theta_{m}}(\tilde{m}_t | r_{\leq t}, \tilde{z}_t), \\
\mathcal{L}_{z} &= -\sum_{t=1}^{l_r} \tilde{m}_{t-1} \cdot \log p_{\theta_{z}}(\tilde{z}_t | r_{<t}, \tilde{z}_{<t}, \tilde{m}_{t-1}).
\label{weak_loss}
\end{aligned}
\vspace{-1mm}
\end{equation}
The learning algorithm is summarized and provided in the appendix.

\begin{table*}[]
\centering
\resizebox{0.9\linewidth}{!}{
\begin{tabular}{llcccccccccccc}
\toprule
\multicolumn{1}{c}{\multirow{2}{*}{Training Data}} & \multicolumn{1}{c}{\multirow{2}{*}{Models}} & \multicolumn{4}{c}{Wizard Seen} & \multicolumn{4}{c}{Wizard Unseen} & \multicolumn{4}{c}{CMU$\_$DoG} \\
\cmidrule(lr){3-6}\cmidrule(lr){7-10}\cmidrule(l){11-14}
\multicolumn{1}{c}{} & \multicolumn{1}{c}{} & PPL & F1   & D-1   & D-2   & PPL & F1   & D-1   & D-2 & PPL   & F1   & D-1   & D-2   \\ \midrule
\multirow{3}{*}{Reddit Corpus} & BART         & 40.1 & 18.4 & 0.076 & 0.355 & 42.9 & 18.4 & 0.049 & 0.237 & 75.8 & 9.8  & 0.021 & 0.131 \\
& ZRKGC        & 41.1 & 18.9 & 0.055 & 0.246 & 42.7 & 18.8 & 0.037 & 0.179 & 53.8 & 12.2 & 0.015 & 0.094 \\ 
& Our Model    & \textbf{35.9} & \textbf{19.3} & \textbf{0.082} & \textbf{0.383} & \textbf{38.4} & \textbf{19.2} & \textbf{0.060} & \textbf{0.292} & 60.4 & 12.2          & \textbf{0.028} & \textbf{0.186} \\ \midrule
\multirow{3}{*}{\begin{tabular}[c]{@{}l@{}}Reddit Corpus + \\ 10\% annotated data\end{tabular}} & BART & 32.7 & 18.9 & 0.073 & 0.357 & 35.0 & 18.8 & 0.049 & 0.235 & 49.5 & 10.1 & 0.019 & 0.110 \\
& ZRKGC        & 29.1 & 19.1    & 0.072  & 0.309 & 31.6 & 18.9 & 0.048 & 0.209 & 38.0 & 13.7 & 0.010 & 0.062 \\ 
& Our Model    & \textbf{28.6} & \textbf{20.4} & 0.073 & \textbf{0.366} & \textbf{30.7} & \textbf{20.0} & 0.052 & \textbf{0.270} & 40.8 & \textbf{14.4} & 0.015 & \textbf{0.122} \\ 
\bottomrule
\end{tabular}
}
\vspace{-2mm}
\caption{Automatic evaluation results. Numbers in bold mean that the improvement to the best performing baseline is statistically significant (t-test with $p$-value $<$ 0.05).}
\vspace{-2mm}
\label{tab:exp_main}
\end{table*}

\section{Experiments}

\subsection{Datasets}

We test our model on benchmarks of knowledge-grounded dialogue generation, including Wizard of Wikipedia (Wizard) \cite{dinan2018wizard} and CMU Document Grounded Conversations (CMU$\_$DoG) \cite{zhou2018dataset}. 
We choose the Reddit Corpus published by \cite{li2020zero} as $\mathcal{D}$ for pre-training. Since it is abundant in expression style as a corpus from online forum, the two latent variables could be well initialized. 
We use part of the training data of Wizard and CMU$\_$DoG as $\mathcal{D}_{sty}$ respectively, for these two datasets are distinctive in expression style and differ from each other. The dialogues in CMU$\_$DoG tend to be causal and short, with most utterances irrelevant to knowledge while the responses in Wizard are usually long and knowledgeable, as some phrases are directly extracted from wiki articles. 

More details about the datasets are described in the appendix.

\begin{table*}[]
\centering
\resizebox{0.9\linewidth}{!}{
\begin{tabular}{lcccccccccc}
\toprule
\multicolumn{1}{c}{\multirow{2}{*}{Models}} & \multicolumn{5}{c}{Wizard Seen} & \multicolumn{5}{c}{CMU$\_$DoG} \\ 
\cmidrule(lr){2-6}\cmidrule(lr){7-11}
\multicolumn{1}{c}{} & Fluency & \begin{tabular}[c]{@{}c@{}}Context\\ Coherence\end{tabular} & \begin{tabular}[c]{@{}c@{}}Knowledge\\ Relevance\end{tabular} & \begin{tabular}[c]{@{}c@{}}Style\\ Consistency\end{tabular} & Kappa & Fluency & \begin{tabular}[c]{@{}c@{}}Context\\ Coherence\end{tabular} & \begin{tabular}[c]{@{}c@{}}Knowledge\\ Relevance\end{tabular} & \begin{tabular}[c]{@{}c@{}}Style\\ Consistency\end{tabular} & Kappa \\ \midrule
BART  & 1.68 & 1.56 & 1.52 & 1.34 & 0.64 & 1.62 & 1.57 & 1.55 & 1.31 & 0.63  \\ 
ZRKGC & 1.62 & 1.59 & 1.55 & 1.36 & 0.65 & 1.61 & 1.53 & 1.65 & 1.56 & 0.66  \\ 
Our Model   & 1.71 & 1.64 & 1.66 & 1.77 & 0.60 & 1.61 & 1.66 & 1.63 & 1.76 & 0.74  \\ 
\bottomrule
\end{tabular}
}
\vspace{-2mm}
\caption{Human evaluation results on learning structure style. }
\vspace{-2mm}
\label{tab:human_structure}
\end{table*}

\subsection{Experimental Setup}
In this paper, we mainly consider two experimental setups, corresponding to the two aspects of knowledge expression. 
To explore how our model can be used to control the distribution of different kinds of segments (knowledge-related and knowledge-irrelevant), we first train the model on the Reddit Corpus and then fine-tune it on a small amount of examples in Wizard and CMU$\_$DoG, respectively.\footnote{We provide the evaluation results of training on the whole Wizard and CMU$\_$DoG in the appendix.}
To verify whether our model can generate the knowledge-related segments in the desired style, we still train the model on the Reddit Corpus, and use a style tag to control the generation process. In this experimental setup, we are primarily concerned with generating with two kinds of styles, positive and negative, where $z_t=2 \cdot \min(1, z_t)$ tells the model to generate a response in positive sentiment and $z_t=3 \cdot \min(1, z_t)$ is for response in negative sentiment.

\paragraph{Evaluation Metrics.}
Following \citet{dinan2018wizard}, we choose PPL and unigram F1 as the metrics to evaluate the appropriateness.
We further use Distinct-$1/2$ (D-$1/2$), which are calculated as ratios of distinct unigrams and bigrams in responses respectively, to evaluate the distinctness.
We also employ classification accuracy as the evaluation metrics for style control experiments.\footnote{We exploit Roberta trained on the SST-2 training set \cite{socher2013recursive} as the evaluator.}
Due to space limitation, we provide automatic evaluation results on more metrics (i.e., BLEU-$1$, METEOR, and ROUGE-L) in the appendix.

To further verify whether our model could learn structure style and content style, we randomly sample $300$ examples from Test Seen of Wizard, and the test set of CMU$\_$DoG respectively, and recruit $6$ well-educated native speakers to do qualitative analysis on the responses generated by our model and all baselines. The annotators need to judge the quality of the responses from four aspects (i.e., \emph{fluency}, \emph{context coherence}, \emph{knowledge relevance} and \emph{style consistency}), and assigns a score from $\{0, 1, 2\}$ (representing ``bad'', ``fair'' and ``good'' respectively) to each response for each aspect.
The agreement among all annotators is measured via Fleiss' kappa \cite{fleiss1971measuring}.
More details about the setup of human evaluation as well as the results on learning content style are provided in the appendix.

\subsection{Baselines}

For the exploration of structure style, we select the following models as baselines: 
(1) \emph{BART} \cite{lewis2020bart}: a model that achieves state-of-the-art performance on various text generation tasks. Note that our model degrades into BART once we remove the module indicator Z and the boundary indicator M; 
(2) \emph{Zero-resource Knowledge-grounded Conversation (ZRKGC)} \cite{li2020zero}\footnote{\scriptsize\url{https://github.com/nlpxucan/ZRKGC}}: a model that is based on UniLM \cite{dong2019unified} and optimized with Generalized EM method. 

For the content style, we consider the following models as baselines: (1) \emph{Emotional Chatting Machine (ECM)} \cite{zhou2018emotional}\footnote{\scriptsize\url{https://github.com/thu-coai/ecm}}: a model which can generate appropriate responses not only content-relevant but also emotional consistent; (2) variant of \emph{DialoGPT} \cite{zhang2019dialogpt}: We add a sentiment indicating token at the first of the sequence and explore whether such simple heuristics works for controlling knowledge expression; (3) \emph{CTRL} \cite{keskar2019ctrl} \footnote{\scriptsize\url{https://github.com/salesforce/ctrl}} : a large-scale model trained on conditional codes to govern the style and content of generation. 

Our model and all baselines are trained on the identical Reddit Corpus to maintain fairness.

\begin{figure*}
  \centering
  \vspace{-2mm}
  \subfigure[{Wizard Seen}] { \label{fig:seen_finetune}
    \includegraphics[width=0.6\columnwidth]{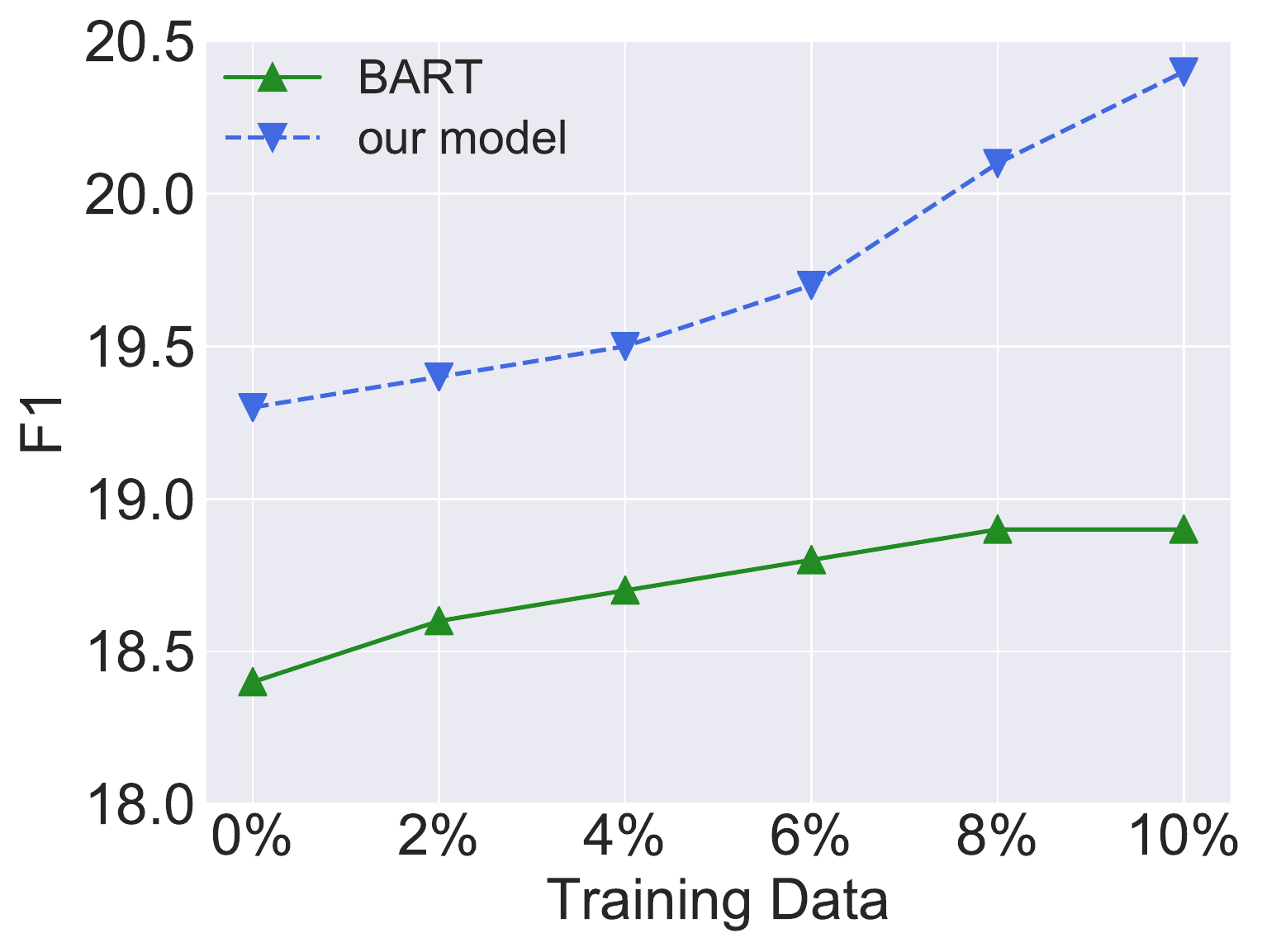}
  }
  \subfigure[{Wizard Unseen}] { \label{fig:unseen_finetune}
    \includegraphics[width=0.6\columnwidth]{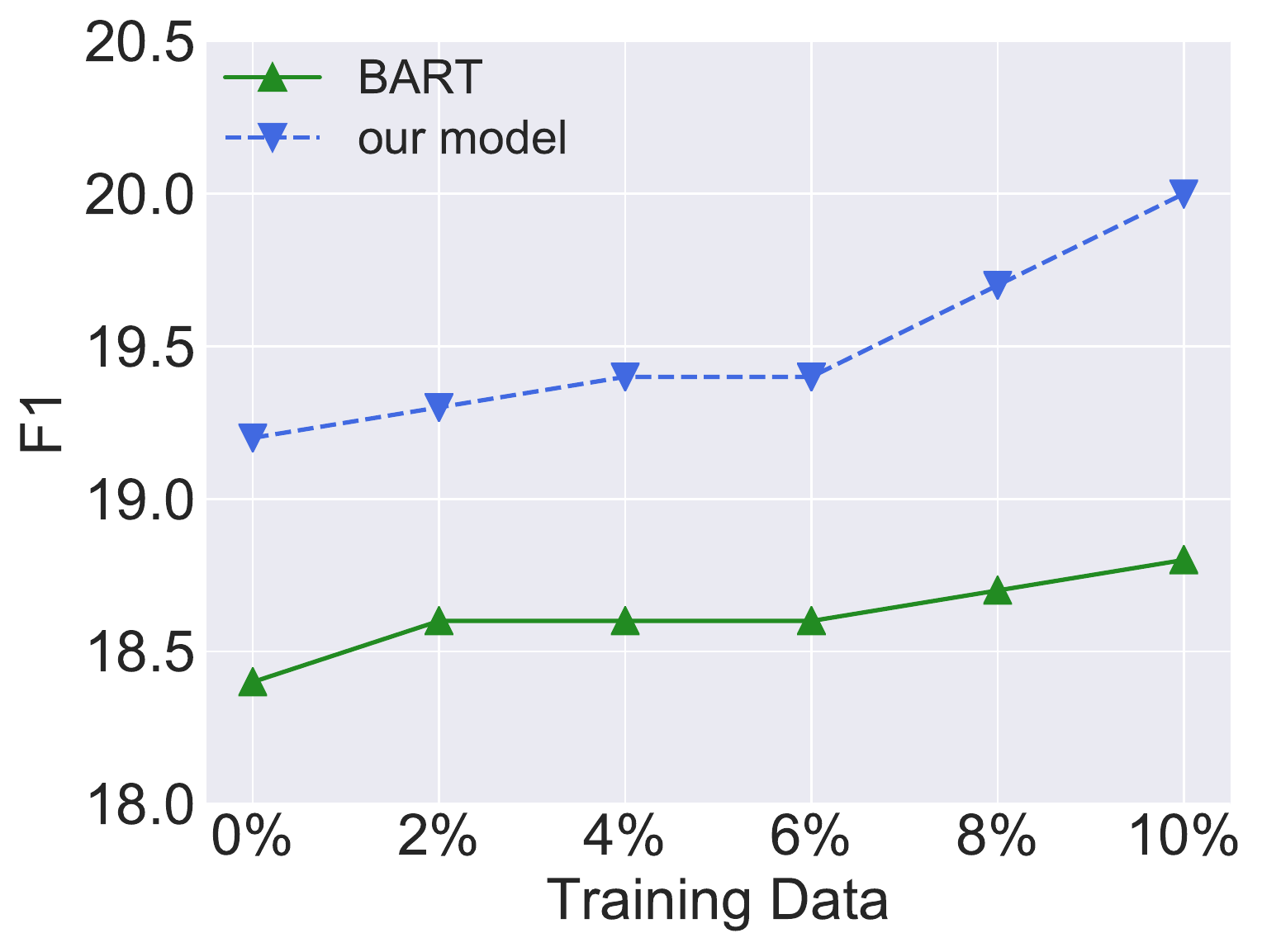}
  }
  \subfigure[{CMU$\_$DoG}] { \label{fig:cmudog_finetune}
    \includegraphics[width=0.6\columnwidth]{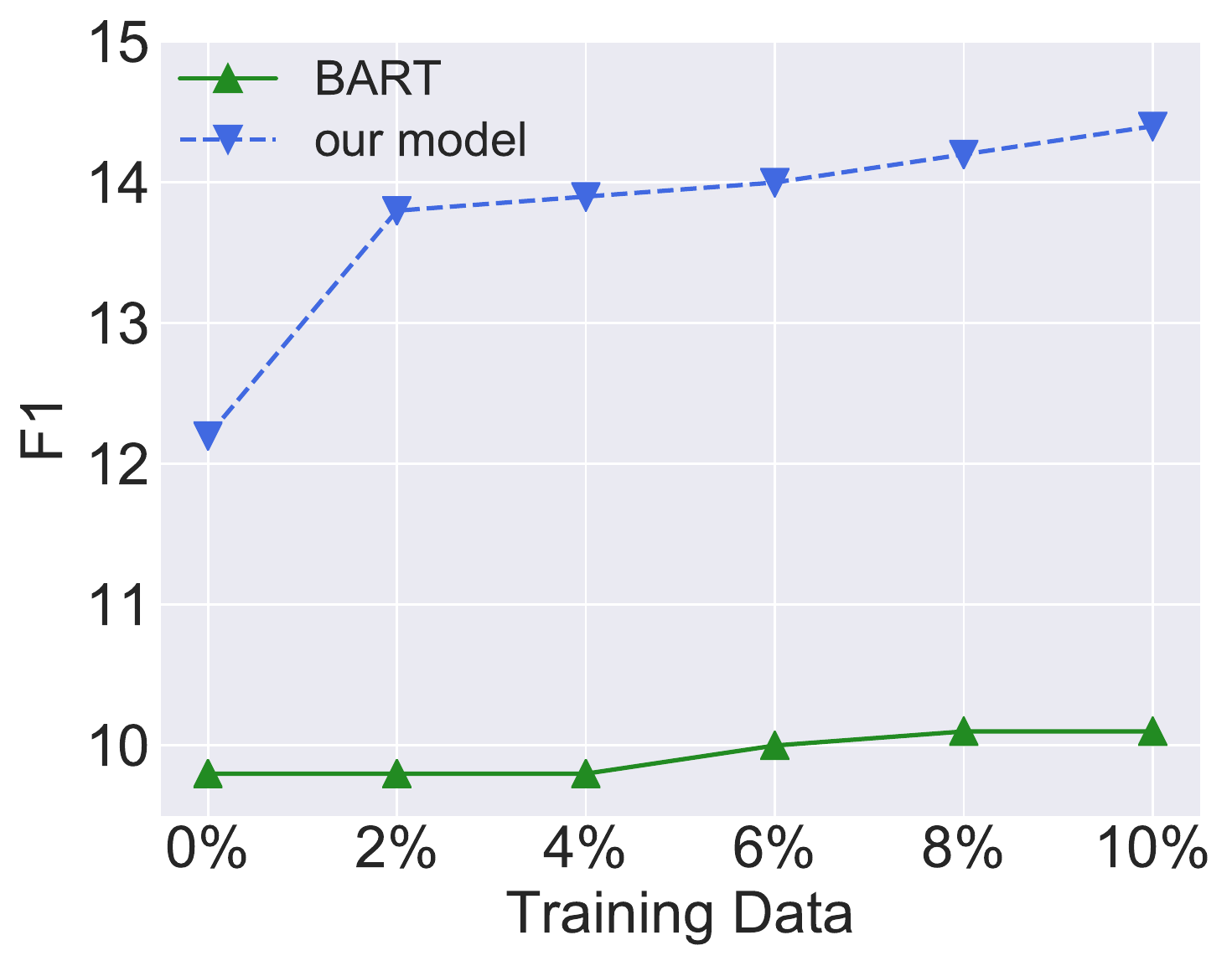}
  }
  \vspace{-4mm}
  \caption{Performance of different models wrt. training data size.}
  \vspace{-1mm}
  \label{fig:dis1}
\end{figure*}

\subsection{Results on Learning Structure Style}
In this section, we demonstrate the effectiveness of our segmentation-based generation framework in both low-resource setting and zero-resource setting and empirically verify that our model can learn structure style with a few annotated examples.

In zero-resource setting, we trained our model on the Reddit Corpus published by \cite{li2020zero} and tested on Wizard and CMU$\_$DoG respectively. Automatic evaluation results are shown in Table \ref{tab:exp_main}.
It could be observed that: 
(1) our model significantly outperforms ZRKGC and BART on most metrics and achieves the new state-of-the-art performance on Wizard. It is impressive that our model exceeds BART in CMU$\_$DoG especially since the proposed model degrades into BART without two sequential latent variables Z and M. The result serves as strong evidence for the effect of two latent variables, which enable the model to learn complex expression style in Reddit Corpus to handle flexible expression in  CMU$\_$DoG. By contrast, BART is far from satisfying with only a regular decoder. 
(2) our model exceeds ZRKGC significantly in terms of Distinct metrics, for ZRKGC mainly focuses on leveraging external knowledge sources for response generation, but falls short on expression diversity. 
In the low-resource setting, after training our model on the Reddit Corpus, we then fine-tune it with only $10\%$ training size of Wizard and CMU$\_$DoG respectively (i.e., $\mathcal{D}_{sty}$ in Sec \ref{sec:overview}) to adjust $p(z_t)$ and $p(m_t)$ to a new structure style. 
When provided with only $10\%$ training data, our model gets obvious improvement ($\sim 1\%$ increase in F1) in contrast with BART ($\sim 0.5\%$ increase in F1) and ZRKGC ($\sim 0.2\%$ increase in F1), proving that the proposed model can learn more sophisticated structure style through quick adjustment on a specific dataset with little cost.\footnote{After trained with 10\% Wizard data, the diversity of our model decreases for it fits the specific expression style of Wizard.}

\begin{table*}[h!]
\centering
\resizebox{0.75\linewidth}{!}{
\begin{tabular}{lcccccccccccc}
\toprule
\multicolumn{1}{c}{\multirow{3}{*}{Models}} & \multicolumn{4}{c}{Wizard Seen}                              & \multicolumn{4}{c}{Wizard Unseen}                            & \multicolumn{4}{c}{CMU$\_$DoG}                               \\ 
\cmidrule(lr){2-5}\cmidrule(lr){6-9}\cmidrule(l){10-13}
\multicolumn{1}{c}{}                        & \multicolumn{2}{c}{positive} & \multicolumn{2}{c}{negative} & \multicolumn{2}{c}{positive} & \multicolumn{2}{c}{negative} & \multicolumn{2}{c}{positive} & \multicolumn{2}{c}{negative} \\ 
\cmidrule(lr){2-3}\cmidrule(lr){4-5}\cmidrule(lr){6-7}\cmidrule(lr){8-9}\cmidrule(lr){10-11}\cmidrule(l){12-13}
 & F1            & Acc           & F1            & Acc           & F1            & Acc           & F1            & Acc           & F1            & Acc           & F1            & Acc           \\  \midrule
ECM                                          & 10.5          & 55.8          & 10.2          & 60.7          & 10.1          & 55.7          & 10.1          & 57.6          & 7.6           & 41.5          & 8.3           & 55.4          \\ 
DialoGPT                                     & 12.1          & 54.1          & 12.1          & 46.9          & 12.0          & 56.0          & 12.0          & 45.0          & 9.2           & 44.9          & 9.2           & 55.1          \\ 
CTRL  &  15.3 & 71.9 & 14.9 & 55.3 & 14.9 & 75.0 & 14.6& 52.3& 9.3 & 70.2 & 9.2 & 61.7 \\
Our Model                                    & \textbf{19.7} & 70.3 & \textbf{19.2} & \textbf{70.7} & \textbf{19.4} & 73.1 & \textbf{19.2} & \textbf{69.9} & \textbf{12.7} & \textbf{74.8} & \textbf{12.2} & \textbf{68.0} \\ 
\bottomrule
\end{tabular}
}
\vspace{-2mm}
\caption{Evaluation results on sentiment control. Numbers in bold mean that the improvement to the best performing baseline is statistically significant (t-test with $p$-value $<$ 0.05).}
\label{tab:exp_senti}
\vspace{-5.5mm}
\end{table*}

\paragraph{Human Evaluation. }
Table \ref{tab:human_structure} shows human evaluation results on learning structure style. It could be observed that: (1) our model is significantly superior to others on \emph{style consistency}, indicating that the model can learn a consistent expression style with very little data. 
(2) our model has better performance on \emph{context coherence} and \emph{knowledge relevance}, tallying with its impressive performance in the low-resource scenario.

\paragraph{Fine-tuning with limited annotated data.}
We first train the model on the Reddit Corpus and then fine-tune it with the amount of annotated data (e.g., Wizard and CMU$\_$DoG) gradually increasing
from $2\%$ to $10\%$. To have a more intuitive understanding of the effects of latent variables Z and M, we compare the proposed model with BART, which generates the response with a single decoder. The evaluation results are shown in Figure \ref{fig:dis1}. It can be concluded from the result that: (1) our model can learn the expression style of a particular dataset more efficiently. As the training data increases, our model has a more significant improvement in terms of the F1 metric; (2) our model performs better in meager resources since there is a considerable gap between our model and BART when the training data is close to $0\%$; (3) the expression style of CMU$\_$DoG can be learned with less data because the model has a significant change in performance after using $2\%$ CMU$\_$DoG training data.

\begin{figure}[!t]
\centering
\includegraphics[width=0.3\textwidth]{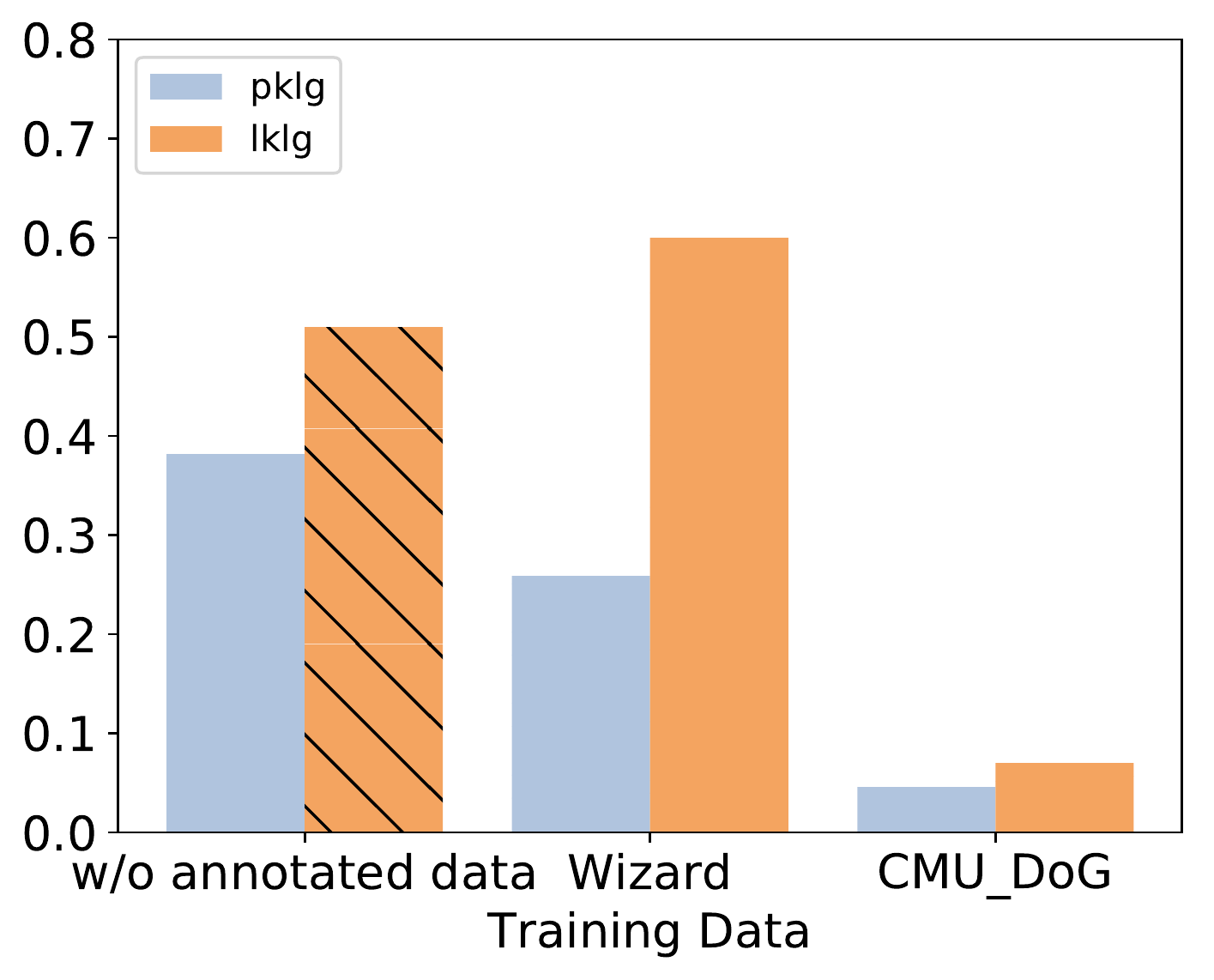}
\vspace{-2mm}
\caption{The effect of fine-tuning on different data.}
\vspace{-5mm}
\label{fig:dis2}
\end{figure}

\paragraph{Refashioning of knowledge-related segments.}
To know how our model adjusts to different datasets, we compare the knowledge-related segments before and after trained with annotated data from two aspects: (1) the average proportion of knowledge-related segments ($pklg$) in a sentence; (2) the average proportion of words belonging to knowledge-related segments ($lklg$). The motivation behind is that the frequency and length of these two kinds of segments generally indicates how well the latent variable is learned to capture the knowledge expression structure. We identify these two kinds of segmentation by comparing their lexical similarities with the knowledge.
Figure \ref{fig:dis2} reports the results. 
The results indicate that our model could learn the underlying structure style of both datasets, with the great difference of $pklg$ and $lklg$ before and after fine-tuning as evidence. 
After fine-tuning with Wizard data, $pklg$ drops to $0.26$ while the $lklg$ grows up a bit, indicating that the knowledge-related segments generated by our model are fewer and longer, which tallies with the fact that the responses in Wizard are probably directly copied from background knowledge. However, after CMU$\_$DoG data is fed to the model, both $pklg$ and $lklg$ shrinks drastically, which agrees with the fact that crowd-sourcing workers converse more liberally online and the responses are less relevant to the background knowledge.

\vspace{-1mm}
\subsection{Results on Learning Content Style}
\vspace{-1mm}
We further investigate whether the proposed model could express knowledge with the desired sentiment. Specifically, we introduce two sets of style adapters to endow knowledge expression in two different sentiments, namely positive and negative. So in this scenario, it is required that responses are not only coherent with context but also limited in positive or negative sentiment.
To apply ECM on knowledge-grounded conversation, we label the sentiment category for each response with a classifier pre-trained on the SST \cite{socher2013recursive} training set.
For DialoGPT, we similarly annotate each response with a sentiment category and append the sentiment token before the context tokens.
The evaluation results is shown in Table \ref{tab:exp_senti}. We can conclude that: (1) The proposed model outperforms all baseline models in terms of all metrics, which indicates that our model can control the sentiment of knowledge expression and guarantee high quality of the generated responses; (2) Simply adding a sentiment indicating token at the beginning of the sequence can not effectively control the style of knowledge expression, as the performance of DialoGPT on sentiment control is poor; (3) Although ECM is designed for sentiment control, it still fails to perform well in this task, proving that sentiment control in the knowledge-grounded conversation is rather difficult. Besides, ECM can only control the sentiment of the whole response but is helpless to manage every knowledge-related segment at a fine-grained level.

\section{Conclusions}

We explore knowledge expression in knowledge-grounded conversation and break down the expression style of a response into the structure of the response (structure style) and the style of the content in each part (content style). We propose a variational segmentation-based generation model to discover the underlying expression style in response. Specifically, we introduce two latent variables to model these two aspects of expression style respectively and induce an evidence lower bound of the likelihood. 
Evaluation results on two benchmarks of the task indicate that our model can learn the structure style with little cost and generate responses in desired content style without any human-annotated data.

\section*{Ethical Considerations}
It's crucial for an open-domain dialogue system to be able to automatically detect and discover the underlying structural pattern of a sentence. With the ability to handle a variety of expression styles, whether positive or negative, serious or casual, our work suggests that we are getting closer to the goal of creating an artificial intelligent dialogue system that can freely communicate with humans, which is beyond the wildest dreams of most AI and NLP researchers. However, a detailed survey should be undertaken in advance to consider the immediate audience's and developers' interests, as well as any potential stakeholder groups.

Furthermore, knowledge-grounded dialogue systems have the potential to fabricate facts and distribute rumors and false information, particularly when the source of external background knowledge is unreliable. If the knowledge candidate set is contaminated by fake news, the response generated by the dialogue system is likely to suffer from the ``hallucination'' issue. Controlling the source of knowledge sentences, such as paragraphs extracted from the wiki, authoritative news sites, or authoritative product documents, is a necessary and practical strategy.

\section*{Acknowledgement}

Thanks for the anonymous reviewers for their constructive comments. This work was supported by the National Key Research and Development Program of China (No. 2020AAA0106600), National Natural Science Foundation of China (NSFC Grant No. 62122089 and No. 61876196), and Beijing Outstanding Young Scientist Program (No. BJJWZYJH012019100020098). Rui Yan is also supported by Beijing Academy of Artificial Intelligence~(BAAI).

\bibliography{anthology}
\bibliographystyle{acl_natbib}

\clearpage
\appendix
\section{Appendix}

\subsection{Derivation of ELBO}

\begin{small}
\begin{equation}
\begin{aligned}
    &\log p(R | U, K) \\
    &= \log \sum_{(M, Z)} p(R, M, Z) \\
    &= \log \sum_{(M, Z)} q(M, Z | R) \frac{p(R, M, Z)}{q(M, Z | R)} \\
    &= \log \mathbb{E}_{(M, Z) \sim q(M, Z | R)} \frac{p(R, M, Z)}{q(M, Z | R)} \\
    &\geq \mathbb{E}_{(M, Z) \sim q(M, Z | R)} \log \frac{p(R, M, Z)}{q(M, Z | R)} \\
    &= \mathbb{E}_{(M, Z) \sim q(M, Z | R)} \log p(R | M, Z) \\
    & - \mathbb{E}_{(M, Z) \sim q(M, Z | R)}\big(\log q(M, Z | R)\\
    & - \log p(M, Z)\big).
\end{aligned}
\end{equation}
\end{small}
According to the mean-filed approximation, $q(M, Z) \approx q(M)q(Z)$. Therefore, $\mathbb{E}_{(M, Z) \sim q(M, Z | R)} \log p(R | M, Z)$ and $\mathbb{E}_{(M, Z) \sim q(M, Z | R)}\big(\log q(M, Z | R) - \log p(M, Z)\big)$ can be re-written as:

\begin{small}
\begin{equation}
\begin{aligned}
 &\mathbb{E}_{(M, Z) \sim q(M, Z | R)} \log p(R | M, Z) \\
 &= \mathbb{E}_{M \sim q(M | R)} \big(\mathbb{E}_{Z \sim q(Z | M, R)} \sum_{t=1}^{l_r}\log p(r_t | r_{<t}, z_t))\big)
\end{aligned}
\end{equation}
\end{small}

\begin{small}
\begin{equation}
\begin{aligned}
 &\mathbb{E}_{(M, Z) \sim q(M, Z | R)}\big(\log q(M, Z | R) - \log p(M, Z)\big) \\
 &=\mathbb{E}_{M \sim q(M | R)}\bigg(\mathbb{E}_{Z \sim q(Z | M, R)}\big(\log q(M | R) -\log p(M)\big) \\
 & + \mathbb{E}_{Z \sim q(Z | M, R)}\big(\log q(Z | M, R) -\log p(Z)\big) \bigg) \\
 &=\mathbb{E}_{M \sim q(M | R)}\bigg(\log q(M | R) -\log p(M)\bigg) \\
 & + \mathbb{E}_{M \sim q(M | R)} \bigg(\mathbb{E}_{Z \sim q(Z | M, R)}\big(\log q(Z | M, R) -\log p(Z)\big)\bigg) \\
 &=\sum_{t=1}^{l_r}\bigg(\mathbb{E}_{M \sim q(M | R)}\big(\log q(m_t) - \log p(m_t)\big)\bigg) \\
 &+ \mathbb{E}_{M \sim q(M | R)}\bigg(\sum_{t=1}^{l_r}m_{t-1} \cdot \mathbb{E}_{Z \sim q(Z | M, R)}\big(\log q(z_t) -\log p(z_t)\big)\bigg) \\
 &=\sum_{t=1}^{l_r}D_{\mathrm{KL}}(q(m_t) \| p(m_t)) \\
 & + \mathbb{E}_{M \sim q(M | R)}\bigg(\sum_{t=1}^{l_r}m_{t-1} \cdot D_{\mathrm{KL}}(q(z_t) \| p(z_t))\bigg).
\end{aligned}
\end{equation}
\end{small}

\subsection{Details about the Construction of $\mathbf{\tilde{M}}$ and $\mathbf{\tilde{Z}}$}
In this section, we provide more details about of construction of $\tilde{M}$ and $\tilde{Z}$. 
For every response in the training set, we parse it as a syntax tree using StanfordNLP toolkit~\cite{manning2014stanford}. The syntax tree we obtain is in a hierarchical and nested structure. The root node of the tree represents the whole response sentence and the root node of every subtree represents a corresponding phrase, a small part of a sentence. For example, if a phrase could be divided into three parts, then the node representing the phrase has three child nodes and each represents a part of the phrase. After we acquire the parsing tree, segmentation is then carried out recursively. To be concrete, we traverse the parsing tree by deep-first search order. Every time we arrive at a node, compute the similarity\footnote{We use unigram Precision to calculate the similarity.} between the knowledge and the phrase represented by the node. If the similarity is above the threshold $\mu_{seg}$, we mark the phrase as a segment and search in this branch terminates. Else we continue to search the child nodes of the current node to segment at a more refined level. We use $\tilde{M}=\{\tilde{m}_t\}_{t=1}^{l_r}$ to denote the results of segmentation labeling.

The pseudo label of module choice $\tilde{Z}=\{z_t\}_{t=1}^{l_r}$ is tagged in a similar way to multiclass classification.
Specifically, for a segment $(r_s, \cdots, r_e)$ where $s$ and $e$ are the start and end position of a  segment respectively. If the similarity between this segment and the knowledge falls below a threshold $\mu_{knl}$, its pseudo label $(z_{s}, \cdots, z_{e})$ will be set to $0$. Otherwise we send the segment to a series of style discriminators one after another until the classification confidence given by a discriminator is above $\mu_{sty_i}$ and pseudo module choice label will be set to $i+1$. If all discriminators fail to classify the segment at a confidence greater than $\mu_{sty_i}$, $(z_{s}, \cdots, z_{e})$ are all $1$, indicating knowledge should be expressed without particular style.

\subsection{Learning Algorithm}
The learning algorithm is summarized in Algorithm \ref{algo}.

\begin{algorithm*}[]
\small
\begin{algorithmic}[1]
    \State {\bfseries Input:} Training data $\mathcal{D}$, thresholds for weak supervision $\mu_{seg}$, $\mu_{knl}$ and $\mu_{sty}$, discriminator$\{Dis_i\}_{i=1}^{N_{sty}}$, maximum step $M$, adapter training step $M^{'}$.
    \For {$m \gets 1 \mbox{~to~} M$}
        \State Sample a mini-batch $\{(U_i, K_i, R_i)\}$ from $\mathcal{D}$.
        \State Conduct segmentation on $R_i$ to get $\tilde{M}$.
        \For {$i \gets 1 \mbox{~to~} N_{seg} $}
            \For {$j \gets 1 \mbox {~to~} N_{sty}$}
            \State Use  $Dis_j$ to classify response segment $(r_{s_i}, \cdots, r_{e_i})$.
            \If {Confidence of $Dis_j \geq \mu_{sty}$ and $(z_{s_i}, \cdots, z_{e_i})$ are not assigned}
            \State $(z_{s_i}, \cdots, z_{e_i}) \gets j+1$
            \EndIf
            \EndFor
        \EndFor
        
        \If {$m \leq M'$}
            \State Update the adapters based on the first term in ELBO.
        \Else
            \State Update the parameters $\theta$ (i.e., $\theta_{m}$, $\theta_{z}$ and the parameters in $p(r_t)$) and $\phi$ (i.e., $\phi_{m}$ and $\phi_{z}$) based on ELBO and Weak Supervision.
        \EndIf
    \EndFor
    \State {\bfseries return}  Generation Model $p_{\theta}(R | U, K)$ with prior distribution $p_{\theta_{m}}$ and $p_{\theta_{z}}$
\end{algorithmic}
\caption{Learning Algorithm}
\label{algo}
\end{algorithm*}

\subsection{Details of Datasets}
\paragraph{Training Data.}
We choose the Reddit Corpus published by~\cite{li2020zero} as $\mathcal{D}$ for pre-training. 
The data contains $842,521$ context-knowledge-response triples for training and $2,737$ context-knowledge-response triples for validation. On average, each dialogue contains $3.1$ utterances in both sets, and the average length of the utterance is $16.0$ in training and is $16.1$ in validation. 

\paragraph{Evaluation Data.}
We test our model on benchmarks of knowledge-grounded dialogue generation, including Wizard of Wikipedia (Wizard)~\cite{dinan2018wizard} and CMU Document Grounded Conversations~(CMU$\_$DoG)~\cite{zhou2018dataset}. 
Both datasets are split into training sets, validation sets, and test sets by the data owners.
We follow~\citet{dinan2018wizard} and conduct the pre-processing with the code published on ParlAI\footnote{\scriptsize\url{https://github.com/facebookresearch/ParlAI/blob/master/projects/wizard\_of\_wikipedia}}. 
Topics in Wizard cover a wide range~($1,365$ in total), and each conversation happens between a wizard who has access to the knowledge about a specific topic and an apprentice who is just eager to learn from the wizard about the topic. The test set is split into two subsets. Test Seen only contains dialogues with topics that have already appeared in the training set, while topics in Test Unseen never appear in the training set and the validation set. 
Different from Wizard, CMU$\_$DoG focuses on movie domain, and besides wizard-apprentice conversations, the data also contain conversations between two workers who know the document and try to discuss the content in depth.
In both datasets, only the turns where knowledge is accessible are considered in response generation. 
Table \ref{tbl:stat} reports the statistics of the Wizard data and the CMU$\_$DoG data

\begin{table*}[]
\centering
\resizebox{0.65\linewidth}{!}{
\begin{tabular}{lccccccc}
\toprule
\multirow{2}{*}{}          & \multicolumn{4}{c}{Wizard of Wikipedia}   & \multicolumn{3}{c}{CMU$\_$DoG} \\ 
\cmidrule(lr){2-5}\cmidrule(lr){6-8}
                           & Train   & Valid  & Test Seen & Test Unseen & Train    & Valid  & Test    \\ \midrule
$\#$ Utterances       & 166,787 & 17,715 & 8,715     & 8,782       & 74,717   & 4,993  & 13,646  \\ \midrule
$\#$ Conversations    & 18,430  & 1,948  & 965       & 968         & 3,373    & 229    & 619     \\ \midrule
$\#$ Topics/Documents & 1,247   & 599    & 533       & 58          & 30       & 30     & 30      \\ \midrule
Avg. $\#$ of Turns & 9.0     & 9.1    & 9.0       & 9.1         & 22.2     & 21.8   & 22.0    \\
\bottomrule
\end{tabular}
}
\caption{Statistics of the Wizard data and the CMU$\_$DoG data.}
\label{tbl:stat}
\end{table*}

\begin{table*}[]
\centering
\resizebox{1.0\linewidth}{!}{
\begin{tabular}{llcccccccccccc}
\toprule
\multirow{2}{*}{Training   Data} & \multicolumn{1}{c}{\multirow{2}{*}{Models}} & \multicolumn{4}{c}{Wizard Seen}  & \multicolumn{4}{c}{Wizard Unseen} & \multicolumn{4}{c}{CMU\_DoG}       \\
\cmidrule(lr){3-6}\cmidrule(lr){7-10}\cmidrule(lr){11-14}
                                 & \multicolumn{1}{c}{}                        & PPL  & BLEU-1 & ROUGE-L & METEOR & PPL   & BLEU-1 & ROUGE-L & METEOR & PPL  & BLEU-1 & ROUGE-L & METEOR-L \\
                                \midrule
\multirow{3}{*}{Reddit Corpus}   & BART                                        & 40.1 & 0.206  & 0.164   & 0.099  & 42.9  & 0.202  & 0.167   & 0.099  & 75.8 & 0.141  & 0.117   & 0.060    \\
                                 & ZRKGC                                       & 41.1 & 0.225  & 0.163   & 0.099  & 42.7  & 0.220  & 0.164   & 0.100  & 53.8 & 0.161  & 0.128   & 0.075    \\
                                 & Ours                                        & 35.9 & 0.235  & 0.168   & 0.095  & 38.4  & 0.232  & 0.169   & 0.095  & 60.4 & 0.166  & 0.131   & 0.069    \\ \midrule
\multirow{3}{*}{\makecell[c]{Reddit Corpus\\+10\% training data }}   & BART                                        & 32.7 & 0.203  & 0.174   & 0.103  & 35.0  & 0.199  & 0.175   & 0.103  & 49.5 & 0.141  & 0.122   & 0.067    \\
                                 & ZRKGC                                       & 29.1 & 0.227  & 0.175   & 0.101  & 31.6  & 0.222  & 0.176   & 0.100  & 38.0 & 0.173  & 0.139   & 0.083    \\
                                 & Ours                                        & 28.6 & 0.237  & 0.181   & 0.103  & 30.7  & 0.231  & 0.176   & 0.101  & 40.8 & 0.182  & 0.139   & 0.080   \\
                                 \bottomrule
\end{tabular}
}
\caption{More Results about Automatic Evaluation.}
\label{tab:exp_more_auto}
\end{table*}

\subsection{More Implementation Details}
We employ a knowledge selection(KS) module to select the top $7$ related sentences in knowledge. The KS module is implemented based on Roberta-base(125M) and trained on the Reddit Corpus. 
Specifically, we treat the sentence which has the highest F1 score with the response as the positive sample, and the negative sample is randomly sampled from all the other knowledge sentences. We train the KS module via maximum likelihood estimation (MLE) with a batch size of $64$ and an initial learning rate of $1e-5$.
The threshold $\mu_{seg}$, $\mu_{knl}$, $\mu_{pos}$ and $\mu_{neg}$\footnote{We consider positive and negative sentiment style in our experiments.} in weak supervision are set as $0.9$, $0.5$, $0.8$ and $0.8$, respectively.
The encoder-decoder architecture is implemented on the basis of Bart-base(139M) and trained on the Reddit Corpus with a batch size of $64$ and an initial learning rate of $5e-6$. 
The parameters for prior and posterior distributions of $Z$ and $M$(i.e., $\theta_{z}$, $\theta_{m}$, $\phi_{z}$ and $\phi_{m}$) are initialized randomly, and optimized with a learning rate of $1e-4$. 
The parameters for adapters are initialized randomly and optimized with a learning rate of $2e-3$. We only train the adapters for the first $1000$ steps.
We utilize gated recurrent units (GRUs) as the basic units in $f_{z-\mathrm{rnn}}$.
We set the hidden size and the number of layers of RNN in our model(i.e., $f_{z-\mathrm{rnn}}$ and $\psi(\cdot)$) as $128$ and $1$ respectively. 
The embedding size for $Z$ is set as $128$ and the adapter size is set as $64$.
When fine-tuning the model on the Wizard and CMU$\_$DoG datasets, the learning rate and the batch size are set as $5e-5$ and $32$ respectively.
We employ greedy search in response decoding. All models are learned with Adam \cite{kingma2014adam} optimizer with $\beta_1=0.9$ and $\beta_2=0.999$. 
We increase the learning rate linearly for the first $200$ steps and decrease it thereafter proportionally to the inverse square root of the step number.  
Early stopping on validation is adopted as a regularization strategy. 
All models are trained on a $8\times$RTX 2080 Ti machine.

\subsection{More Results about Automatic Evaluation}

Table \ref{tab:exp_more_auto} reports more results about the automatic evaluation, from which we can see that our model still outperforms the baselines.

\subsection{Human Evaluation}

We randomly sample $300$ examples from Test Seen of Wizard, and the test set of CMU$\_$DoG respectively, and recruit $6$ well-educated native speakers to do qualitative analysis on the responses generated by our model and all baselines. 
For each of the $300$ examples, an annotator is provided with the context, the ground-truth knowledge, model responses and the associated style types. 
For evaluation of structure style, we defined two kinds of structure styles based on two datasets, namely the Wizard-like style $S_{wizard}$ and the CMU$\_$DoG-like style $S_{cmudog}$. While for evaluation of content style, we roughly divide content styles in two categories, $S_{pos}$ and $S_{neg}$ for convenience. The responses provided by different models are randomly shuffled to hide their sources. 
The annotators need to judge the quality of the responses from four aspects: (1) \emph{fluency}: whether the response is fluent without any grammatical errors; (2) \emph{context coherence}: whether the response is coherent with the context; (3) \emph{knowledge relevance}: whether the response is relevant with the knowledge; and (4) \emph{style consistency}: whether the response exhibits the desired style. Each annotator assigns a score from $\{0, 1, 2\}$ (representing ``bad'', ``fair'' and ``good'' respectively) to each response for each aspect. Each response obtains four scores for aforementioned four aspects, and the agreement among all annotators is measured via Fleiss' kappa \cite{fleiss1971measuring}.

\begin{table*}[]
\centering
\resizebox{0.9\linewidth}{!}{
\begin{tabular}{lcccccccccc}
\toprule
\multicolumn{1}{c}{\multirow{2}{*}{Models}} & \multicolumn{5}{c}{Wizard Seen} & \multicolumn{5}{c}{CMU$\_$DoG} \\ 
\cmidrule(lr){2-6}\cmidrule(lr){7-11}
\multicolumn{1}{c}{} & Fluency & \begin{tabular}[c]{@{}c@{}}Context\\ Coherence\end{tabular} & \begin{tabular}[c]{@{}c@{}}Knowledge\\ Relevance\end{tabular} & \begin{tabular}[c]{@{}c@{}}Style\\ Consistency\end{tabular} & Kappa & Fluency & \begin{tabular}[c]{@{}c@{}}Context\\ Coherence\end{tabular} & \begin{tabular}[c]{@{}c@{}}Knowledge\\ Relevance\end{tabular} & \begin{tabular}[c]{@{}c@{}}Style\\ Consistency\end{tabular} & Kappa \\ \midrule
ECM      & 0.85 & 0.94 & 1.02 & 1.24 & 0.65 & 0.96 & 0.95 & 1.18 & 1.08 & 0.72  \\
DialoGPT & 1.57 & 1.41 & 1.19 & 1.26 & 0.75 & 1.55 & 1.62 & 1.09 & 1.02 & 0.65  \\
Our      & 1.64 & 1.60 & 1.78 & 1.72 & 0.76 & 1.59 & 1.63 & 1.51 & 1.69 & 0.62  \\ 
\bottomrule
\end{tabular}
}
\caption{Human evaluation results on learning content style.}
\label{tab:human_content}
\end{table*}

\paragraph{Results on Learning content style.}

Table \ref{tab:human_content} reports the human evaluation results on learning content style. The three models are trained on the Reddit Corpus. We can conclude that: (1) by introducing two latent variables and a number of adapters for different styles, our model can generate responses in desired content style (i.e., $S_{pos}$ and $S_{neg}$) more accurately and achieve significant improvement on \emph{style consistency}, which is consistent with the results in Table \ref{tab:exp_senti}; 
(2) our model also outperforms ECM and DialoGPT on \emph{fluency}, \emph{context coherency} and \emph{knowledge relevance} thanks to the capacity of large-scale pre-trained language models and the introduction of external knowledge respectively.

\begin{table*}[]
\centering
\resizebox{0.7\linewidth}{!}{
\begin{tabular}{llccccccccc}
\thickhline
\multicolumn{1}{c}{\multirow{2}{*}{Training Data}} & \multicolumn{1}{c}{\multirow{2}{*}{Models}} & \multicolumn{3}{c}{Wizard Seen} & \multicolumn{3}{c}{Wizard Unseen} & \multicolumn{3}{c}{CMU$\_$DoG} \\ 
\cmidrule(lr){3-5}\cmidrule(lr){6-8}\cmidrule(l){9-11}
\multicolumn{1}{c}{} & \multicolumn{1}{c}{} & F1   & D-1   & D-2   & F1   & D-1   & D-2   & F1   & D-1   & D-2    \\ \midrule
\multirow{4}{*}{$100\%$ annotated data} & TMN & 15.9 & 0.041 & 0.176 & 14.3 & 0.025 & 0.106 & 9.9  & 0.003 & 0.008  \\ 
                                & SKT         & 19.3 & 0.085 & 0.300 & 16.1 & 0.056 & 0.188 & -    & -     & -      \\ 
                                & DRD         & 19.3 & 0.065 & 0.252 & 17.9 & 0.046 & 0.177 & 10.7 & 0.010 & 0.044 \\ 
                                & KnowledGPT  & 22.0 & 0.141 & 0.431 & 20.5 & 0.094 & 0.290 & 13.5 & 0.023 & 0.113  \\ 
                                & Our Model & 22.0 & 0.128 & 0.415 & 20.8 & 0.090 & 0.278 & 15.3 & 0.031 & 0.121  \\ 
                                \midrule
\multirow{3}{*}{Reddit Corpus}  & BART        & 18.4 & 0.076 & 0.355 & 18.4 & 0.049 & 0.237 & 9.8  & 0.021 & 0.131  \\ 
                                & ZRKGC       & 18.9 & 0.055 & 0.246 & 18.8 & 0.037 & 0.179 & 12.2 & 0.015 & 0.094  \\ 
                                & Our Model   & 19.3 & 0.082 & 0.383 & 19.2 & 0.060 & 0.292 & 12.2 & 0.028 & 0.186  \\ 
                                \midrule
\begin{tabular}[c]{@{}l@{}}Reddit Corpus + \\ $10\%$ annotated data\end{tabular} & Our Model & 20.4 & 0.073 & 0.366 & 20.0 & 0.052 & 0.270 & 14.4 & 0.015 & 0.122  \\ \midrule
\begin{tabular}[c]{@{}l@{}}Reddit Corpus + \\ $100\%$ annotated data\end{tabular} & Our Model & 21.9 & 0.134 & 0.453 & 21.2 & 0.103 & 0.302 & 15.5 & 0.041 & 0.134  \\ 
\bottomrule
\end{tabular}
}
\caption{Automatic evaluation results.}
\label{tab:exp_more}
\end{table*}

\subsection{Comparison with More Baselines}

We compare with models trained on full training data, and Table \ref{tab:exp_more} shows the evaluation results. First, it is noted that our model outperforms KnowledGPT in terms of F1 by using only $10\%$ training data\footnote{The $10\%$ training data is randomly sampled. The result is an average value of three repetitive experiments on every dataset} on CMU$\_$DoG, which provides a strong support for the effectiveness of the proposed model. Second, by adjusting the structure style on a small amount of data, the gap between our model and KnowledGPT is further narrowed, while the improvement on ZRKGC and BART is trivial.

\subsection{Ablation over Weak Supervision}
\begin{table*}[]
\centering
\resizebox{0.9\linewidth}{!}{
\begin{tabular}{llccccccccc}
\toprule
\multicolumn{1}{c}{\multirow{2}{*}{Training Data}} & \multicolumn{1}{c}{\multirow{2}{*}{Models}} & \multicolumn{3}{c}{Wizard Seen} & \multicolumn{3}{c}{Wizard Unseen} & \multicolumn{3}{c}{CMU$\_$DoG} \\ \cmidrule(lr){3-5}\cmidrule(lr){6-8}\cmidrule(lr){9-11}
\multicolumn{1}{c}{} & \multicolumn{1}{c}{}                 & F1   & D-1   & D-2   & F1   & D-1   & D-2   & F1   & D-1   & D-2      \\ \midrule
\multirow{3}{*}{Reddit Corpus} & Our model                    & 19.3 & 0.082 & 0.383 & 19.2 & 0.060 & \textbf{0.292} & \textbf{12.2} & 0.028 & \textbf{0.186}  \\
                               & -weak supervision on Z       & 19.1 & 0.077 & 0.362 & 19.1 & 0.056 & 0.270 & 10.2 & 0.027 & 0.155    \\ 
                               & -weak supervision on Z and M & 19.1 & 0.083 & 0.382 & 18.8 & 0.058 & 0.270 & 9.5  & 0.023 & 0.147    \\ \midrule
\multirow{3}{*}{\begin{tabular}[c]{@{}l@{}}Reddit Corpus + \\ $10\%$ annotated data\end{tabular}} & Our model & \textbf{20.4} & 0.073 & 0.366 & \textbf{20.0} & 0.052 & \textbf{0.270} & \textbf{14.4} & 0.015 & \textbf{0.122}    \\ 
                               & -weak supervision on Z       & 19.5 & 0.072 & 0.354 & 19.3 & 0.051 & 0.250 & 13.2 & 0.014 & 0.115    \\ 
                               & -weak supervision on Z and M & 19.5 & 0.077 & 0.366 & 19.2 & 0.054 & 0.258 & 13.5 & 0.013 & 0.091    \\ 
\bottomrule
\end{tabular}
}
\caption{Ablation study over the weak supervision. Numbers in bold means that the improvement to variants is statistically significant (t-test with $p$-value $<0.05$)}
\label{tab:ablation}
\end{table*}

To have more insights into the impact of weak supervision on the performance of our model, we compare the proposed model with the following variants: (1)\textit{-weak supervision on Z}: the weak supervision on module indicator Z is removed; (2)\textit{-weak supervision on Z and M}: the weak supervision on module indicator and boundary indicator is removed. Table \ref{tab:ablation} reports the evaluation results. We can conclude that (1) the weak supervision objectives significantly improve model performance; (2) the weak supervision objectives play a more crucial role on CMU$\_$DoG, as removing them causes a dramatic drop in performance. The reason is that this dataset has more sophisticated expression styles and it is difficult to learn these styles without auxiliary supervision signals.

\subsection{Ablation over Boundary Indicator}

\begin{figure}[!t]
\centering
\includegraphics[width=0.3\textwidth]{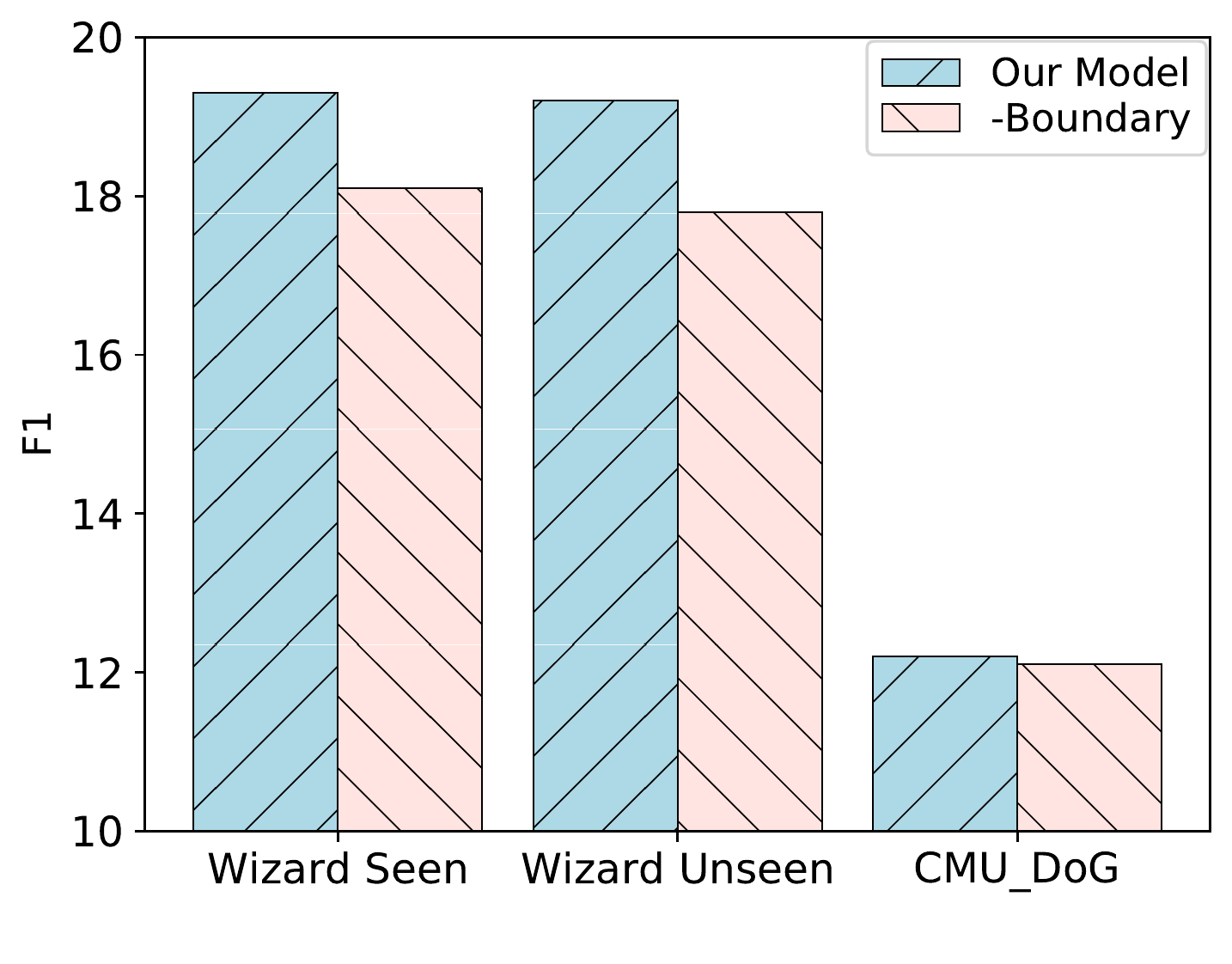}
\vspace{-2mm}
\caption{Ablation results on Wizard and CMU$\_$DoG.}
\label{fig:boundary_ablation}
\vspace{-5mm}
\end{figure}

Since the module indicator is conditioned on the boundary indicator, we are curious about what will happen if the $M$ is removed. The ablation result is shown on Figure~\ref{fig:boundary_ablation}. There is an evident drop on Wizard Seen and Wizard Unseen, verifying the effect of boundary indicator in assisting the module indicator. The margin is tiny on CMU\_DoG, perhaps because its structure feature is easier to capture, so the module indicator could works properly itself.

\subsection{Case Study}

\begin{table*}[h!]
\centering
\includegraphics[width=1.0\textwidth]{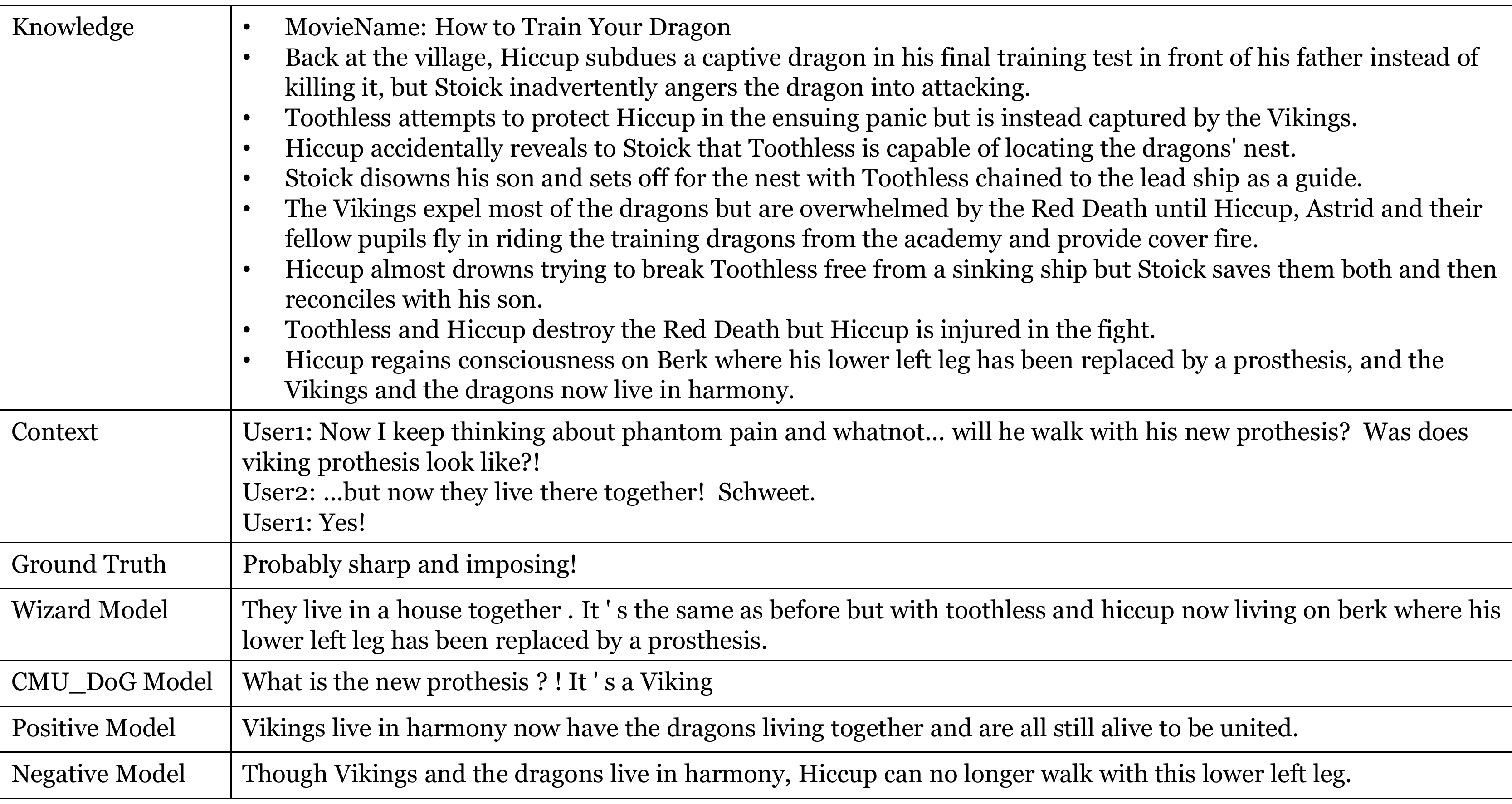}
\caption{A case from test set of CMU$\_$DoG.}
\label{tab:case}
\end{table*}

This section mainly studies how different models vary in knowledge expression for the same context and background knowledge.
Table \ref{tab:case} shows an example from the test set of CMU$\_$DoG. This example contains the background knowledge which gives a plot from the movie, and the dialogue context which is generated by discussing the content in the knowledge. 
We choose the following four models to generate the response in corresponding style given the dialogue context and knowledge, and all models are pre-trained with the Reddit Corpus: (1) Wizard Model for $S_{wizard}$: the model fine-tuned with $10\%$ training data in Wizard; (2) CMU$\_$DoG Model for $S_{cmudog}$: the model fine-tuned with $10\%$ training data in CMU$\_$DoG; (3) Positive Model for $S_{pos}$: the model forced to express knowledge with positive sentiment; (4) Negative Model for $S_{neg}$: the model forced to express knowledge with negative sentiment. 
We can see that the knowlege expression style of the Wizard Model and CMU$\_$DoG Model are quite different. The central part of the Wizard Model response is copied from the background knowledge, which is consistent with the style of Wizard data. The response generated by CMU$\_$DoG Model is more casual in knowledge expression, and the content is mainly related to the conversation context. Besides, responses generated by the Positive Model exhibit evident positive sentiment, while responses generated by the Negative Model 
show relatively negative sentiment.



\end{document}